\documentclass[conference]{IEEEtran}
\def\IEEEtitletopspace{18pt}
\IEEEoverridecommandlockouts
\usepackage{cite}
\usepackage{amsmath,amssymb,amsfonts}
\usepackage{algorithmic}
\usepackage{graphicx}
\usepackage{textcomp}
\usepackage{xcolor}
\usepackage{hyperref}
\usepackage{orcidlink}
\usepackage{graphicx}
\usepackage{booktabs}
\usepackage{caption}
\usepackage{graphicx}
\usepackage{multirow}
\usepackage{booktabs}

\def\BibTeX{{\rm B\kern-.05em{\sc i\kern-.025em b}\kern-.08em
    T\kern-.1667em\lower.7ex\hbox{E}\kern-.125emX}}
\usepackage{balance}

\begin{document}

\title{Cooperverse: A Mobile-Edge-Cloud Framework for Universal Cooperative Perception with Mixed Connectivity and Automation
}

\author{
Zhengwei Bai$^{\orcidlink{0000-0002-4867-021X}}$,~\IEEEmembership{Member, IEEE},
Guoyuan Wu$^{\orcidlink{0000-0001-6707-6366}}$,~\IEEEmembership{Senior Member, IEEE},
Matthew J. Barth$^{\orcidlink{0000-0002-4735-5859}}$,~\IEEEmembership{Fellow, IEEE},\\
Yongkang Liu,
Emrah~Akin~Sisbot,
Kentaro Oguchi

\thanks{Zhengwei Bai, Guoyuan Wu, and Matthew J. Barth are with the Department of Electrical and Computer Engineering, the University of California at Riverside, Riverside, CA 92507 USA (e-mail:  zbai012@ucr.edu).}

\thanks{Yongkang Liu, Emrah~Akin~Sisbot, and Kentaro Oguchi are with Toyota Motor North America, InfoTech Labs, Mountain View, CA 94043, USA.}

}

\maketitle

\begin{abstract}


Cooperative perception (CP) is attracting increasing attention and is regarded as the core foundation to support cooperative driving automation, a potential key solution to addressing the safety, mobility, and sustainability issues of contemporary transportation systems. However, current research on CP is still at the beginning stages where a systematic problem formulation of CP is still missing, acting as the essential guideline of the system design of a CP system under real-world situations. In this paper, we formulate a universal CP system into an optimization problem and a mobile-edge-cloud framework called \textit{Cooperverse}. This system addresses CP in a mixed connectivity and automation environment. A Dynamic Feature Sharing (DFS) methodology is introduced to support this CP system under certain constraints and a Random Priority Filtering (RPF) method is proposed to conduct DFS with high performance. Experiments have been conducted based on a high-fidelity CP platform, and the results show that the \textit{Cooperverse} framework is effective for dynamic node engagement and the proposed DFS methodology can improve system CP performance by 14.5\% and the RPF method can reduce the communication cost for mobile nodes by 90\% with only 1.7\% drop for average precision.
\end{abstract}

\begin{IEEEkeywords}
Cooperative Perception, Mobile-Edge-Cloud Framework, Object Detection, Internet of Things, Mixed Connectivity and Automation
\end{IEEEkeywords}

\section{Introduction}


Technology advancements in intelligent vehicles have led to tremendous improvements in safety, mobility, and sustainability~\cite{bai2022hybrid} in current transportation systems. Empowered by artificial intelligence (AI), automated vehicles (AVs) are achieving remarkable capabilities of perceiving the surrounding environment, laying the foundation of driving automation~\cite{badue2021self}.


In real-world traffic environments, sensor occlusion is one of the most challenging bottlenecks that limit the potential of AVs. The Internet of Things (IoT) breaks the limits mentioned above by sharing information via vehicular communication~\cite{khayyam2020artificial}. Connected vehicles (CVs) can be aware of the statuses of surrounding vehicles or infrastructure(i.e., \textit{Cooperative Awareness}), such as Traffic Signal Phase and Timing (SPaT) information, by utilizing vehicle-to-vehicle (V2V) and vehicle-to-infrastructure (V2I) communications ~\cite{8031051}. However, in mixed traffic conditions, it is more challenging to enable cooperative awareness due to the presence of legacy vehicles without connectivity. 


To cooperatively perceive the environment via spatially separated nodes is critical to address the issues above, defined as Cooperative Perception (CP)~\cite{bai2022survey}. CP relies on connectivity and automation to perceive the environment in a cooperative way~\cite{wang2020v2vnet}. Furthermore, enabled by V2I communication, the sensing performance of CP methods can be improved by leveraging roadside sensors~\cite{9827461}. 


Due to the innate advantage of having complete perception in mixed traffic, CP research has quickly emerged in recent years ~\cite{xu2022v2x, F-cooper, bai2022vinet}. Most CP methods are designed from the theoretical perspective of purely improving the detection performance while omitting the practical constraints that cannot be circumvented in real-world conditions, such as computational power, and communication bandwidth.~\cite{lou2022cooperative}. Additionally, CP system design is primarily considered on a case-by-case basis~\cite{bai2022survey}, and a systematic problem formulation of CP that can support the design of CP systems under universal traffic conditions is still missing. 

Therefore, in this paper, we treat the CP system design as an optimization problem by considering the computational complexity, communication capacity, fusion scheme, and information topology. A mobile-edge-cloud (MEC) framework is proposed by following the problem formulation. A novel feature-sharing methodology is proposed to enable dynamic cooperation under different system constraints. A CP environment is created for experiments and evaluation. 



\section{Related Work}
\label{related work}
\subsection{Vehicle-based object perception}
Due to a significant amount of interest in automated vehicle R\&D over the past decade, onboard object perception techniques have made considerable progress. There have been various algorithms proposed for computer vision based on the sensors used (such as monocular/stereo cameras, LiDAR) or perception tasks (e.g., traffic sign recognition, lane detection, road user detection)~\cite{marti2019review}. The use of convolutional neural networks (CNNs) for camera-based solutions has been widely investigated in recent years, and they also inspire the design of perception pipelines for analyzing point cloud data (PCD) from onboard LiDAR sensors~\cite{lang2019pointpillars}. A combination of cameras and LiDARs can be used to significantly improve ego-vehicle centric perception performance, in which objects can be detected and tracked using the LiDAR, while the targets can be classified using both cameras and LiDARs~\cite{liu2022bevfusion}. 

\subsection{Infrastructure-based object perception}
In order to support more efficient automated driving in a mixed traffic environment, infrastructure-based surveillance systems can provide additional object-level information to target road users beyond traditional data collection (e.g., volume, point speed) based on loop detectors and radar~\cite{bai2022infrastructure}. Hao et al., designed a bottom-up pipeline for infrastructure-based object perception using traditional model-based methods~\cite{zhao2019detection}. Utilizing data-driven models, Bai et al. demonstrated the concept of \textit{Cyber Mobility Mirror} where a roadside LiDAR was used to enable real-time 3D object perception at an intersection~\cite{bai2022cmm}.

\subsection{V2X-based cooperative object perception}
By leveraging both vehicle-based perception and infrastructure-based perception, vehicle-to-everything (V2X) based cooperative object perception is considered to be the most promising pathway towards tapping the full potential of Cooperative Driving Automation (CDA)~\cite{bai2022survey}. Xu et al.~\cite{xu2022v2x} proposed a V2X-based cooperative perception methods considering the heterogeneity of vehicle and infrastructure nodes and multi-scale receptive fields. Lou et al.~\cite{lou2022cooperative} conducted the Proof-of-Concept of CP in real-world by applying V2X to enable entities sharing their sensing results and the program demonstrates CP system can significantly improve the perception capability of the involved entities.


\section{Methodology}
\label{methodology}
\subsection{Problem Formulation for CP System Design}
\label{formulation}
Cooperative perception system design can be formulated as a graph-based constrained optimization programming (COP), where the objective is to maximize the system-wide perception performance metric, $\mathcal{J}(\mathbf{x},\mathbf{y})$, with constraints on the computational resource, $c_{i} \in \mathcal{C}$ of perception node $i, (i \in \mathcal{N})$, and the communication bandwidth, $b_{ij} \in \mathcal{B}$, between node $i$ and node $j$. There are two sets of decision variables. The first set of them, $x_{i} \in \mathbf{x}$, are defined to determine the allocation of information processing resources for each node, i.e., $\varepsilon = \{0, 1, 2\}$ where 0, 1 and 2 represent “raw data preservation” with the computation cost $r_{0}$, “feature extraction” with the computational cost $r_{1}$ and “bounding box determination” with the computational cost $r_{2}$, respectively. The other set of them, $y_{ij} \in \mathbf{y}$, is defined to identify if there is data/information flow routing from node $i$ to node $j$, i.e.,
\begin{equation}
y_{ij} = \left\{\begin{matrix}
1, & \text{if the data from node $i$ to node $j$}\\ 
0, & \text{otherwise} 
\end{matrix}\right.
\end{equation}

Therefore, the CP system topology design problem can be written as follows:
\begin{equation}
\underset{\mathbf{x}}{\max}\mathcal{J}(\mathbf{x},\mathbf{y})
\end{equation}
subject to
\begin{equation}
\begin{matrix}
R(x_{i}) \leq c_{i},& \forall i \in \mathcal{N} \\ 
D(y_{ij}) \leq b_{ij}, & \forall i,j \in \mathcal{N}\\ 
x_{i} = \{0, 1, 2\}, & \forall i \in \mathcal{N}\\
y_{ij} = \left\{\begin{matrix}
1, & \text{if data flows from node $i$ to node $j$}\\ 
0, & \text{otherwise} 
\end{matrix}\right., & \forall i,j \in \mathcal{N}
\end{matrix}
\label{eq: constraints}
\end{equation}
where $R(x_{i})$ denotes the computational effort at node $i$ which depends on not only the deployed fusion scheme (i.e., early fusion, immediate fusion, late fusion), but also the amount of data/information (including both ego-vehicle sensor number/type and incoming flows from other nodes); $D(y_{ij})$ represents the amount of data/information flows node $i$ to node $j$. Please note that in real-world implementation, all the parameters and graph structures can be time-varying.

\subsection{MEC-Enabled CP Framework}

\begin{figure}[!ht]
    \centering
\includegraphics[width=0.49\textwidth]{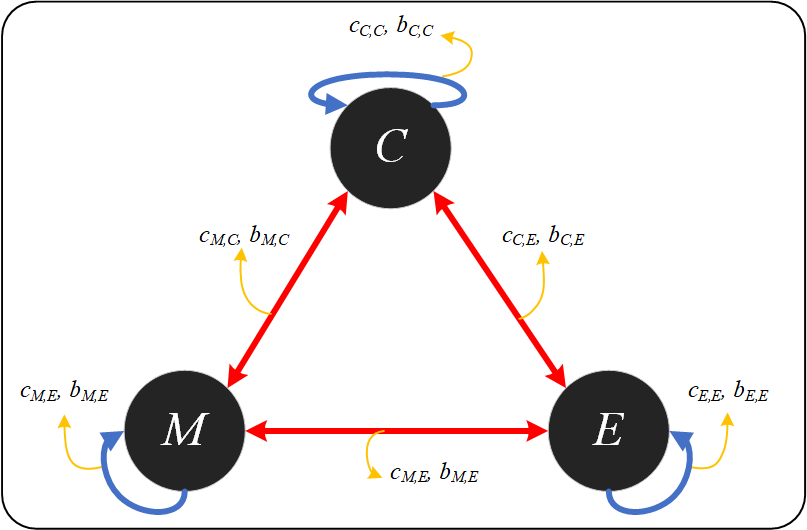}
    \caption{MEC-based CP system framework.}
    \label{fig: mec-cp}
\end{figure}

Based on the problem formulation of CP system design, we propose a mobile-edge-cloud (MEC) framework for designing the CP system from the perspective of real-world implementation. Figure~\ref{fig: mec-cp} illustrates the systematic diagram of the MEC-based CP framework. 

Three types of nodes are designed, including mobile nodes -- the road users, edge nodes -- the infrastructure node, and the cloud node -- the central computing server node. Specifically, in the real-world environment, mobile nodes include Cooperative Perception Vehicles (CPV) which can actively perceive the surrounding environment and cooperate with others via intermediate fusion, CV which have connectivity but can only share their own information due to the lack of environmental sensing capabilities, and CAV which can sense the surrounding condition and share the object-list information via V2X communication. Infrastructure-based perception and communication systems are regarded as edge nodes. Taking advantage of the spatial locations of roadside sensors, edge nodes have a better field of view~\cite{bai2022infrastructure}. Meanwhile, compared with cloud communication, edge-computing systems can achieve lower latency requirements for some time-critical tasks, e.g., collision warnings. But for the large-scale implementation of CP systems, cloud computing is necessary to provide the capacity of dealing with a huge amount of data processing and storage, as well as wide-range message distribution. In general, mobile, edge, and cloud nodes can cooperate with each other under certain constraints formulated in Eq.~\ref{eq: constraints}. 

\begin{figure}[!ht]
    \centering
    \includegraphics[width=0.49\textwidth]{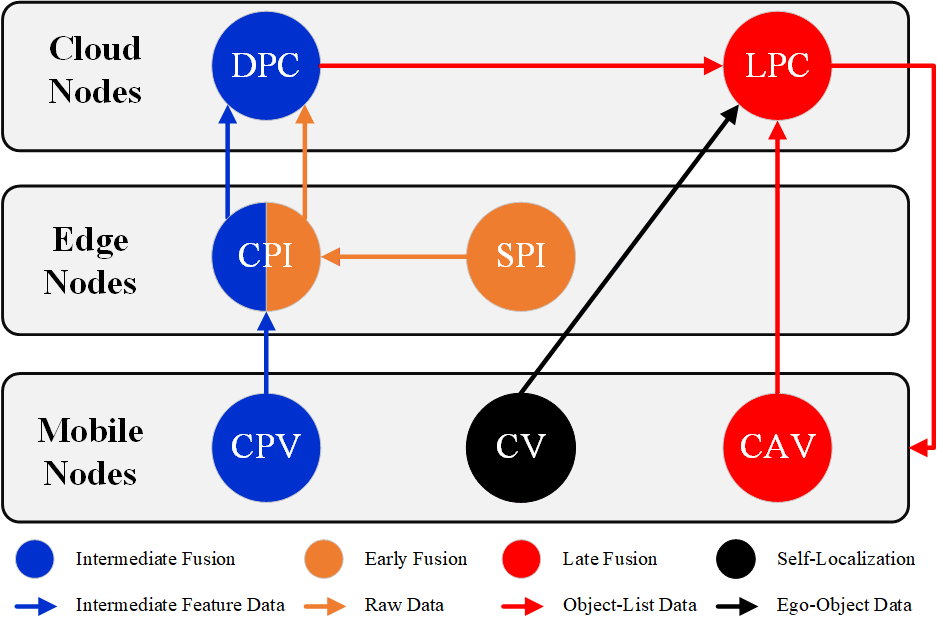}
    \caption{A specification for MEC-based CP system framework.}
    \label{fig: mec-cp-spe}
\end{figure}

In this paper, we investigate the CP system performance by following the MEC framework. The system design is based on the properties of different perception nodes. Figure~\ref{fig: mec-cp-spe} is a specification of the MEC-based CP system framework.

Several core principles are considered when designing the CP system structure above:
\begin{itemize}
    \item Communication constraints: data transmission for mobile nodes are designed for object data or intermediate data, which requires lower bandwidth than raw data. Taking advantage of high-speed wire connection, raw data transmission is feasible in communication between edge nodes and cloud nodes.
    \item Computational constraints: high-end computers are needed only for CAV nodes, Central Perception Infrastructure (CPI) nodes, and Deep Perception Cloud (DPC) nodes, while other nodes in Figure~\ref{fig: mec-cp-spe} only require less computational power.
    \item Information flow complexity: bidirectional information flow is avoided as it will lead to a $\mathcal{O}(N^2)$ communication complexity. In this paper, a forward-backward information flow is designed for $\mathcal{O}(N)$ communication complexity.
    \item Fusion capacity: early fusion, intermediate fusion, and late fusion are combined to support the ad-hoc visitation of different types of nodes.
\end{itemize}

\subsection{Dynamic Feature Sharing}
To address the limitation caused by the communication bandwidth, a dynamic feature sharing (DFS) methodology is proposed by sifting the feature data for sharing based on the specific communication bandwidth. For instance, to fit the lower communication bandwidth, the number of feature cells shared with others needs to be reduced. 

\begin{figure}[!ht]
    \centering
    \includegraphics[width=0.5\textwidth]{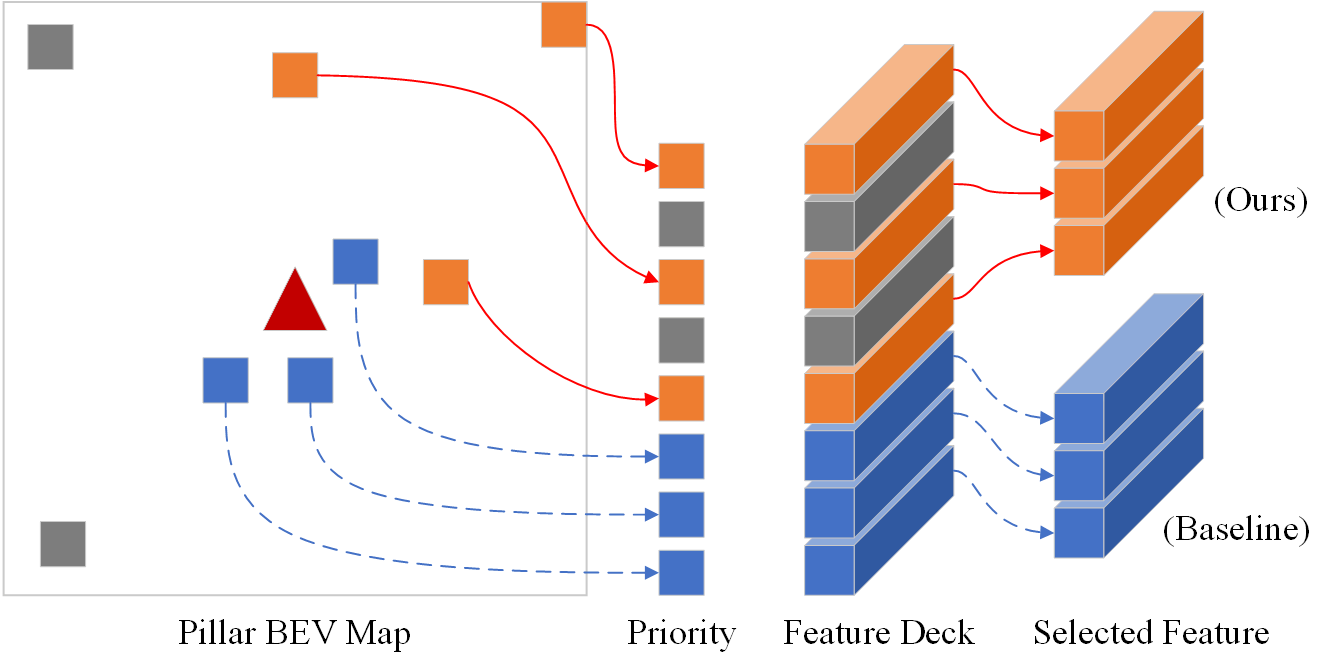}
    \caption{The diagram to illustrate the idea of PFF.}
    \label{fig:dfs}
\end{figure}

In this paper, a Random Priority Filtering (RPF) method is proposed to sift the feature from the Feature Deck (entire feature cells). A visualization diagram is shown in Figure~\ref{fig:dfs} to illustrate the idea of RPF and the difference with the baseline of taking the Top-$\mathcal{K}$ highest priorities. 

As shown in Figure~\ref{fig:dfs}, the triangle represents the spatial location of the sensor and squares represent the voxelized feature cell. Instead of truncating the Feature Deck based on the highest priorities (e.g., $\mathcal{K} = 3$ in Figure~\ref{fig:dfs}), RPF will select the feature with Random-$\mathcal{K}$ priorities. 

The priority calculation is based on the distance between the spatial location of the feature cell and the sensor. For computational efficiency, Manhattan distance is applied for calculating the priorities, as shown below:
\begin{equation}
    \mathcal{D}_{m} = \mid x_{p} - x_{s}\mid + \mid y_{p} - y_{s}\mid
\end{equation}
where $\mathcal{D}_{m}$ is the Manhattan distance between the feature location $(x_{p}, y_{p})$ and the sensor location $(x_{s}, y_{s})$.

\subsection{Dynamic Feature Fusion}
For a universal CP framework, the Cooperverse framework integrates three types of fusion schemes, including early fusion, intermediate fusion, and late fusion.
\subsubsection{Early Fusion}
Based on the MEC structure, the Global Referencing Coordinate (GRC) system is designed for raw data preprocessing as well as early fusion. Specifically, raw sensor data will be transformed into a static GRC to align all the information together.  

\begin{equation}
\mathcal{P}^{E\rightarrow G} = 
\begin{bmatrix}
\mathcal{R}_{X}& 0\\
0&1
\end{bmatrix}
\cdot
\begin{bmatrix}
\mathcal{R}_{Y}& 0\\
0&1
\end{bmatrix}
\cdot
\begin{bmatrix}
\mathcal{R}_{Z}& 0\\
0&1
\end{bmatrix}
\cdot\mathcal{P}^{E} + \mathcal{T}^{E\rightarrow G}
\end{equation}

\noindent where $\mathcal{R}_{X}$, $\mathcal{R}_{Y}$, $\mathcal{R}_{Z}$, and  $\mathcal{T}^{E\rightarrow G}$ represent the rotation matrix along $x-$axis, $y-$axis, $z-$axis, and the translation matrix from ego-coordinate to GRC, respectively. $\mathcal{P}^{S}$ and $\mathcal{P}^{S\rightarrow G}$ represent the raw data before and after the transformation. 

Then the early fusion will be applied as follows:
\begin{equation}
    \mathcal{P}_{early} = \bigcup_{i=1}^{n} \mathcal{P}^{E\rightarrow G}_{i}
\end{equation}
where $n$ represents the number of nodes involved in the early fusion process.

\subsubsection{Intermediate Fusion}
\begin{figure}[!ht]
    \centering
    \includegraphics[width=0.5\textwidth]{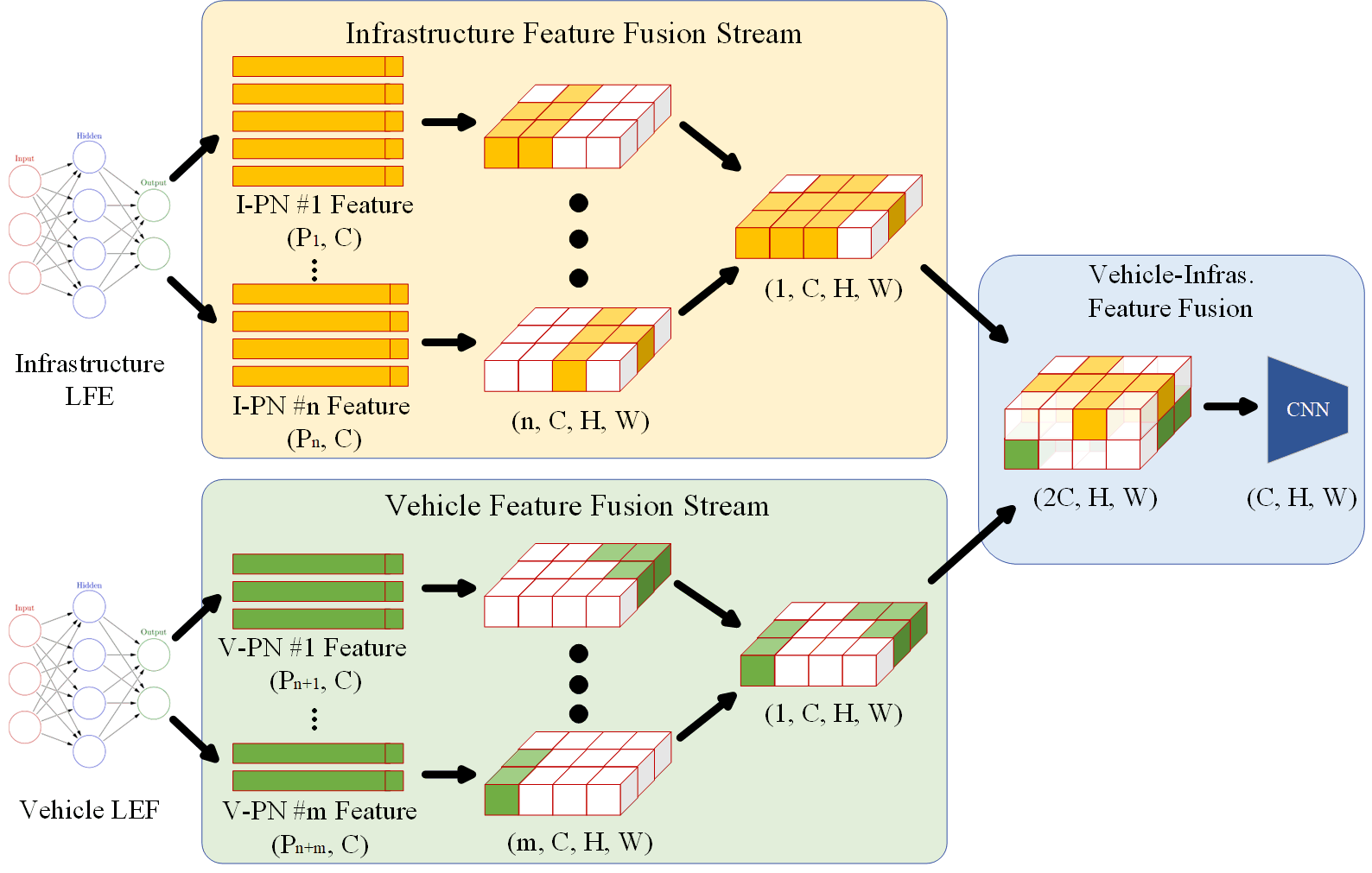}
    \caption{Diagram for Two-Stream Fusion process.}
    \label{fig: tsf}
\end{figure}

For supporting Intermediate Fusion under dynamic, scalable, and heterogeneous conditions, VINet~\cite{bai2022vinet} is adopted as the basic neural network to generate the universal feature map for intermediate fusion-based nodes. The Two-Stream Fusion in VINet is demonstrated in Fig~\ref{fig: tsf} in which features from vehicle nodes and infrastructure nodes are extracted and fused.

To be specific, RFF is applied to generate sifted features which will be fed into TSF for cooperative fusion. 

\subsubsection{Late Fusion}
For late fusion, Non-Maximum Suppression (NMS) is designed in the Cooperverse framework. Specifically, a multi-class circle-NMS is adopted to filter the bounding boxes from multiple multi-class object lists.


\section{Experiments}
\label{experiments}

\subsection{Dataset Acquisition}

``\textit{CARTI}'' (i.e., \textbf{CAR}la-ki\textbf{T}t\textbf{I}) platform is applied for collecting the LiDAR sensor data and ground truth labels for CP system model training and testing. Specifically, two infrastructure nodes and three vehicle nodes are deployed for data collection. The sensor configuration in the CARTI platform is based on the experience from a real-world perception system~\cite{bai2022cmm} which is shown in Table~\ref{tab:parameter}. Totally $8,695$ frames of 3D point clouds are collected, including $4,347$ frames for training, $1,449$ frames for evaluation, and $2,899$ frames for testing. 

\begin{table}[!ht]
  \centering
  \caption{Parameter Configuration (onboard/roadside) in the CARTI Platform}
  \resizebox{0.35\textwidth}{!}{%
    \begin{tabular}{c|c}
    \toprule
    Parameters  & Setting  \\
    \midrule
    LiDAR Channels  & $64$ \\
    LiDAR Height  & $1.74/4.74m $ \\
    LiDAR Sensing Range  & $100.0m $ \\
    LiDAR Rotation Frequency  & $10.0$Hz\\
    Upper FOV  & $+22.5~/~+0$ \\
    Lower FOV  & $-22.5~/~-22.5$ \\
    Standard Deviation of Points   & $0.01m$ \\
    General points Dropoff rate & $45\%$ \\
    General points Dropoff intensity & $0.8$ \\
    \bottomrule
    \end{tabular}}%
  \label{tab:parameter}%
\end{table}%


\subsection{Experimental Setup}
\subsubsection{Perception Environment}
\begin{figure}[!ht]
    \centering
    \includegraphics[width=0.5\textwidth]{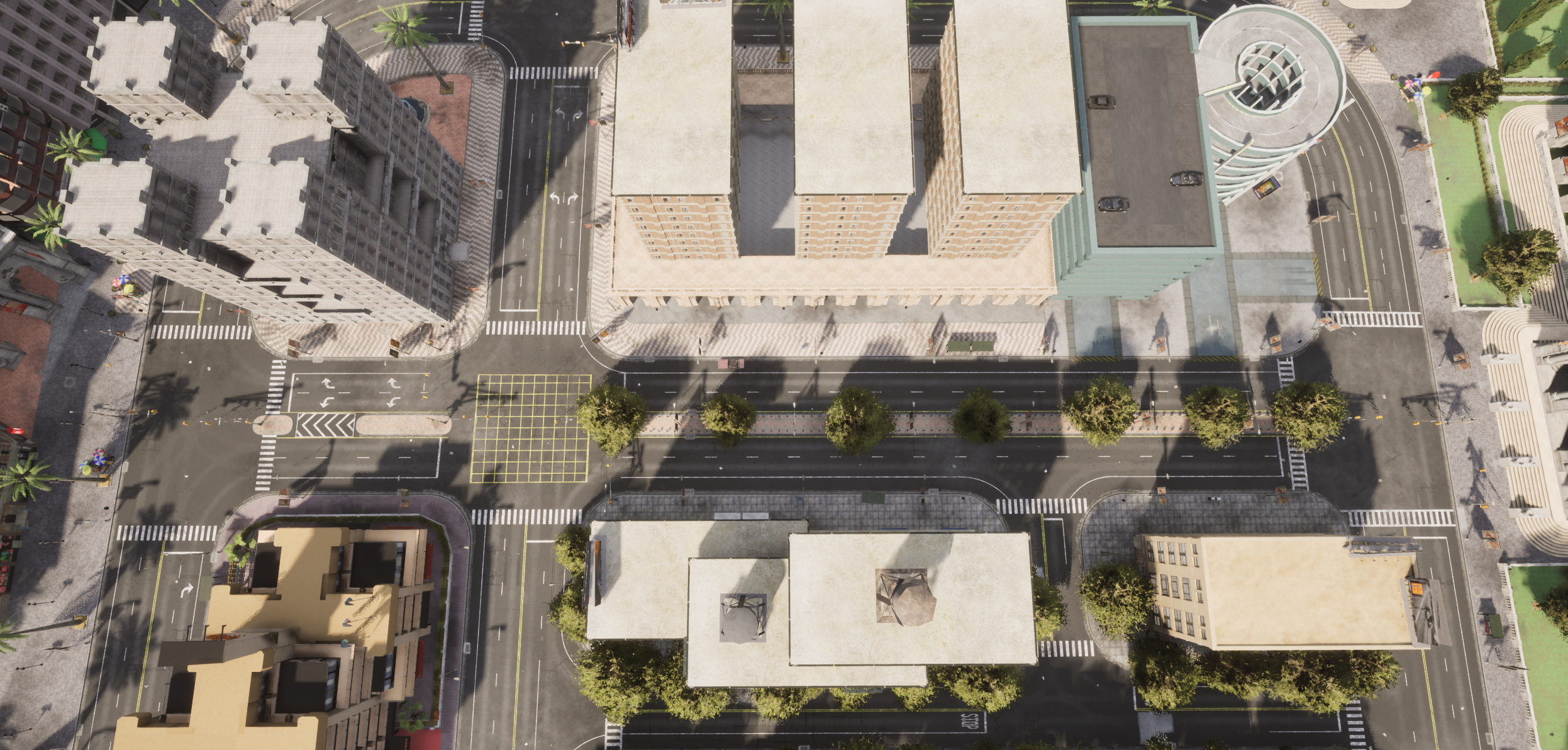}
    \caption{Top-down view of the CP environment.}
    \label{fig:scene}
\end{figure}

To evaluate and investigate the performance of the Cooperverse framework, a three-adjacent intersection network with a size of $280m\times80m$ is coded, as shown in Figure~\ref{fig:scene}

\subsubsection{Training Details}
The training and testing platform consists of an Intel$^\text{\textregistered}$ Core™ i7-10700K CPU and an NVIDIA RTX 3090 GPU. The training pipeline is designed with 160 epochs with \textit{Batchsize} of 2. The voxel size is set as [$0.23m, 0.23m, 8.00m$] and the maximum number of voxels per node $\mathcal{K}$ is set as $45,000$ for training and various values decreasing from  $15,000$ to $500$ for testing the performance under different constraints.

\begin{table*}[!ht]
\centering
\caption{AP performance under different node combinations.}
\resizebox{0.78\linewidth}{!}{%
\begin{tabular}{c|c||c||c|c|c||c|c|c} 
\toprule
\multicolumn{2}{c||}{CP Nodes}                            & \multicolumn{7}{c}{AP}                                                                                      \\ 
\hline
\multirow{2}{*}{Edge Node} & \multirow{2}{*}{Mobile Node} & \multirow{2}{*}{Overall AP} & \multicolumn{3}{c||}{Pedestrians IoU@0.25} & \multicolumn{3}{c}{Car IoU@0.7}  \\ 
\cline{4-9}
                           &                              &                             & MP$\geq$10  & MP$\geq$5   & MP$\geq$1                        & MP$\geq$10  & MP$\geq$5   & MP$\geq$1              \\ 
\midrule
Infra.@1                   & Vehi.@0                      & 18.82                       & 2.44  & 9.44  & 9.54                       & 32.32 & 30.31 & 28.86            \\
Infra.@1                   & Vehi.@1                      & 22.43                       & 4.01  & 5.57  & 5.73                       & 42.29 & 38.50 & 38.49            \\
Infra.@1                   & Vehi.@2                      & 25.75                       & 6.04  & 7.15  & 7.22                       & 44.77 & 44.68 & 44.61            \\
Infra.@1                   & Vehi.@3                      & 29.95                       & 10.73 & 11.94 & 12.06                      & 48.29 & 48.32 & 48.33            \\
Infra.@2                   & Vehi.@0                      & 31.42                       & 2.06  & 9.59  & 10.05                      & 55.91 & 55.49 & 55.39            \\
Infra.@2                   & Vehi.@1                      & 36.11                       & 3.72  & 5.64  & 7.46                       & 66.64 & 66.64 & 66.58            \\
Infra.@2                   & Vehi.@2                      & 39.43                       & 5.89  & 7.41  & 8.79                       & 71.67 & 71.51 & 71.31            \\
Infra.@2                   & Vehi.@3                      & 42.21                       & 10.78 & 12.21 & 13.61                      & 72.28 & 72.23 & 72.16            \\
\bottomrule
\end{tabular}
}
\label{tab:eva-node}
\end{table*}

\subsubsection{Evaluation Details}
The detection performance is measured with Average Precision (AP) and Average Recall (AR) at Intersection-over-Union (IoU) thresholds of 0.25 for pedestrians and 0.7 for cars, respectively. Furthermore, based on the Minimum number of Points (MP) reflected by the ground target, each evaluation class is further divided into three categories: MP$\geq$10, MP$\geq$5, and MP$\geq$1, respectively, to investigate the performance of CP methods on different difficulty levels.

\subsection{Evaluation and Analysis}
In this section, we evaluated different CP systems from three perspectives: 1) various node combinations for cooperative perception, 2) perception performance under different system constraints using DFS, and 3) the trade-off between bandwidth saving and AP reduction using RPF. Due to limited space, system evaluation on dynamic feature fusion will only include intermediate fusion. The performance of a combination of multiple fusion schemes will be further discussed in future work.

\subsubsection{Dynamic Node Engagement}
In this paper, we evaluate the AP performance for different node engagement conditions. It is noteworthy that the model is only trained once under all available nodes engaged, and then evaluated under different conditions without further fine-tuning to make testing situations more realistic.

The testing results are shown in Table~\ref{tab:eva-node}. There are three findings based on analyzing the performance data:
\begin{itemize}
    \item \textbf{\textit{Scalable effectiveness}}: the system performance of object detection, to the most extent, increases when there is a new node engaged in the CP system.
    \item \textbf{\textit{Different significance}}: in the CP system, the importance of different types of nodes is different. According to Table~\ref{tab:eva-node}, infrastructure nodes show much higher significance than vehicle nodes for improving the AP performance. For instance, a CP system with 2 infrastructure nodes achieves higher overall AP than the CP performance under the condition of 1 infrastructure node and 3 vehicle nodes.
    \item \textbf{\textit{Impact heterogeneity}}: Different types of nodes show heterogeneous impacts on specific object classes. For instance, infrastructure nodes have a primary influence on the detection of cars while vehicle nodes show dominant impacts on detecting pedestrians. Since the infrastructure nodes have less occlusion resulting a better car detection. The point cloud information for pedestrians from vehicle nodes can be more sufficient since vehicle nodes are spatially closer to the pedestrians compared with infrastructure nodes.
\end{itemize}

\subsubsection{System-level Sharing Condition}
As mentioned in Section~\ref{formulation}, the CP system can operate under certain constraints. In this paper, we evaluate the CP system under different system-level communication constraints. Taking advantage of the attribute of lightweight feature extracted by VINet, the constraint of communication bandwidth can be linearly reflected in the maximum allowed feature cells for sharing. Considering that each lightweight feature cell requires 256 bytes for transmission~\cite{bai2022vinet}, we tested the system-level total transmission feature cells from $30,000$ to $75,000$, i.e., around $7.7$MB to $19.2$MB per system frame.

\begin{figure}[!t]
    \centering
    \includegraphics[width=0.48\textwidth]{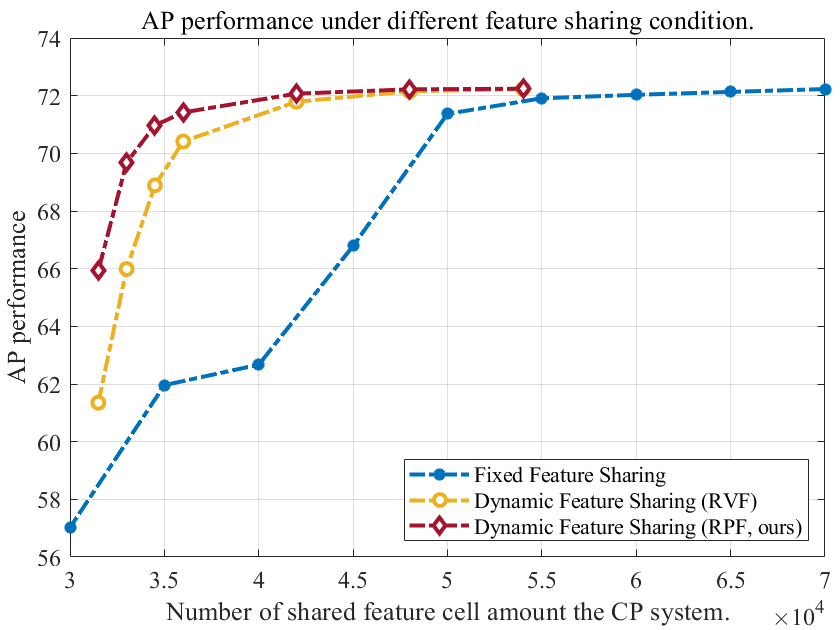}
    \caption{AP performance under different feature sharing conditions.}
    \label{fig:eval-ap-band}
\end{figure}

According to Figure~\ref{fig:eval-ap-band}, the traditional fixed feature sharing method shows a significant performance drop when the system sharing cell goes less than $50,000$. However, our proposed Dynamic Feature Sharing methodology can keep the AP performance until the system-level sharing cells decrease to $35,000$.
Specifically, under the system communication constraint of $8.9$MB per frame, our method can improve AP performance by 14.5\% compared with fixed feature sharing.

\begin{table*}[!t]
\centering
\caption{Detection performance under different feature sharing constraints.}
\resizebox{0.75\linewidth}{!}{%
\begin{tabular}{c|c|c|c|c|c|c|c} 
\toprule
\# of feature sharing     & 8,000 & 6,000 & 4,000 & 2,000 & 1,500 & 1,000 & 500    \\ 
\hline
Top-K nearest Filtering   & 72.18 & 71.46 & 66.32 & 61.7  & 61.57 & 61.49 & 61.24  \\
Top-K Farest Filtering    & 72.22 & 72.01 & 71.58 & 70.34 & 68.21 & 66.29 & 61.71  \\
Random Voxel Filtering    & 72.23 & 72.15 & 71.79 & 70.41 & 68.89 & 65.99 & 61.35  \\
Random Priority Filtering (Ours) & 72.24 & 72.22 & 72.07 & 71.42 & 70.97 & 69.68 & 65.93  \\
\bottomrule
\end{tabular}
}
\label{tab:dfs}
\end{table*}

\subsubsection{Trade-off for Dynamic Feature Sharing}
In this section, we investigate the performance of different DFS methods. Since infrastructure nodes have a higher impact on the system-level performance and mobile nodes tend to have a lower communications bandwidth, we fixed the $\mathcal{K}$ of infrastructure nodes to $15,000$ while exploring the AP performance under different constraints of vehicle nodes and the evaluation results are shown in Table~\ref{tab:dfs}. 

\begin{figure}[!t]
    \centering
    \includegraphics[width=0.5\textwidth]{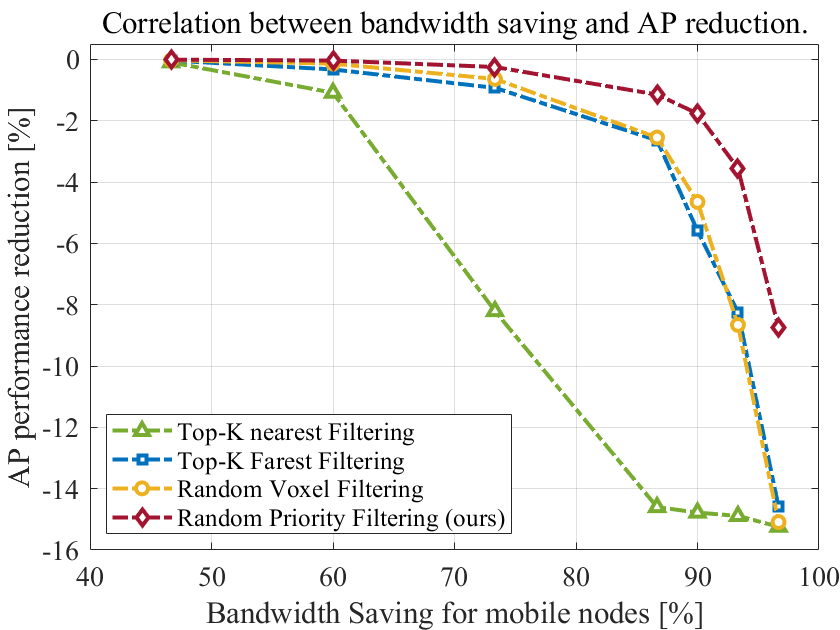}
    \caption{Correlation between bandwidth saving and AP reduction.}
    \label{fig:eval-ap-corr}
\end{figure}

Figure~\ref{fig:eval-ap-corr} demonstrates the correlation between bandwidth saving and AP reduction. Our RPF method has the highest resistance to performance dropping when a higher bandwidth constraint applies. Specifically, RPF can reduce 90.0\% communication bandwidth with only 1.7\% AP reduction compared with fixed feature sharing CP system. Furthermore, in terms of the AP dropping resistance, RPF can surpass the other DFS methods by 51.1\%.

\section{Conclusion\&Future Work}
\label{conclusion}
In this paper, a MEC framework, named  \textit{Cooperverse}, is proposed for supporting universal cooperative perception (CP). An optimization problem is formulated to mathematically investigate the topology design of the CP system. A \textit{Dynamic Feature Sharing} methodology is introduced to support the CP system under certain constraints and a \textit{Random Priority Filtering} method is proposed to conduct DFS with high performance. Experiments have been conducted based on a high-fidelity CP platform, and the results show that the Cooperverse framework is effective for dynamic node engagement and the proposed DFS methodology can improve system perception performance by 14.5\% and the RPF method can reduce the communication cost for mobile nodes by 90\% with only a drop of 1.7\% in AP.

\section*{Acknowledgments}
This research was funded by Toyota Motor North America, InfoTech Labs. The contents of this paper reflect the views of the authors, who are responsible for the facts and the accuracy of the data presented herein. The contents do not necessarily reflect the official views of Toyota Motor North America.

\bibliographystyle{IEEEtran}  
\bibliography{references}  

\end{document}